\documentclass[a4paper]{article}

\usepackage{INTERSPEECH2022}

\usepackage[table]{xcolor}
\usepackage{multirow}
\usepackage{adjustbox}

\title{Multilingual Simultaneous Speech Translation}
\name{Shashank Subramanya$^1$, Jan Niehues$^{1,2}$}
\address{
  $^1$Department of Data Science and Knowledge Engineering, Maastricht University\\
  $^2$Institute for Anthropomatics and Robotics, Karlsruhe Institute of Technology}
\email{shashank.subramanya@student.maastrichtuniversity.nl,jan.niehues@kit.edu}

\begin{document}

\maketitle
\begin{abstract}
  Applications designed for simultaneous speech translation during events such as conferences or meetings need to balance quality and lag while displaying translated text to deliver a good user experience. One common approach to building online spoken language translation systems is by leveraging models built for offline speech translation. Based on a technique to adapt end-to-end monolingual models, we investigate multilingual models and different architectures (end-to-end and cascade) on the ability to perform online speech translation. On the multilingual TEDx corpus, we show that the approach generalizes to different architectures. We see similar gains in latency reduction (40\% relative) across languages and architectures. However, the end-to-end architecture leads to smaller translation quality losses after adapting to the online model. Furthermore, the approach even scales to zero-shot directions.
\end{abstract}

\section{Introduction}
\label{sec:introduction}
    Recent advances in speech translation (ST) has been driven by Transformer based models \cite{pham2019very}. It has led to end-to-end (E2E) systems that have delivered state-of-the-art performance on several benchmarks \cite{ansari2020findings}. In  certain instances, cascaded systems that use automatic speech recognition (ASR) followed by a machine translation (MT) model, have delivered a better performance when optimized by techniques such as back translation  \cite{zhao2021volctrans}. Speech translation also benefits from multilingual models which enable knowledge transfer across various languages and alleviate the issue of data scarcity. Moreover, they allow zero-shot translation across language pairs without training data \cite{li2020multilingual}.

In spoken language translation, there is often a need to produce translations simultaneously, without waiting for the speaker to finish. Offline ST systems are unsuited for the task as they are trained on full sequences and try to generate complete sentences for partial inputs \cite{niehues2018low}. Fine-tuning with partial input sequences followed by implementing a decoding strategy that outputs translated text by comparing consecutive output sequences and displaying parts that agree can be utilized to adapt offline systems for online ST \cite{liu2020low}. This is an efficient technique as compared to re-training from scratch it requires lesser time and resources. Moreover, it is achieved without task-specific external training data while retaining the model’s ability to perform offline ST. Hence, we explore if the adaptation procedure can be utilized to build multilingual online ST models with various architectures. 

The key findings from our experiments are:
\begin{itemize}
    \item Cascaded and end-to-end offline models can be transformed to online models using the same technique. However, cascaded models suffer relatively larger losses in translation quality 
    \item In multilingual models, there is similar improvement in latency (40\% relative) for different languages in both cascade and end-to-end systems post adaptation
    \item The adaptation to streaming inputs generalized to zero-shot directions not seen in training 
\end{itemize}

\section{Related work}
\label{sec:related_work}
\textbf{Multilingual ST} Previous work on multilingual systems included constructing a cascaded pipeline of (mono or multilingual) ASR and (bi or multilingual) MT models \cite{dessloch2018kit}. Advances in sequence-to-sequence modeling led to end-to-end systems that outperformed bilingual models \cite{inaguma2019multilingual}. Multilingual ST models have been further enhanced by using multi-modal data inputs and multi-task (ASR, ST, and MT) learning \cite{zeng2021multilingual}.\\

\noindent \textbf{Simultaneous ST} Low latency ST has evolved from statistical methods \cite{fujita2013simple,sridhar2013segmentation} to neural models \cite{gu2016learning,arivazhagan2019monotonic}. Recent work has shown that a better quality latency trade-off can be achieved by utilizing an RNN Transducer based architecture \cite{liu2021ustc}. Research also focuses on decoding strategies. A prominent approach to dealing with partial input sequences is streaming where the decoder appends the output to a growing hypothesis as new inputs are available \cite{cho2016can,ma2018stacl}. Among the various streaming strategies, the local agreement of hypotheses is shown to have a very good accuracy latency trade-off \cite{liu2020low}.

\section{Offline to online speech translation}
\label{sec:offline_to_online}
Offline ST systems generate a translation only after a speech sentence is completed. To reduce latency, one approach to online ST is to ingest inputs in fixed-size chunks and perform chunk-level decoding. In this approach, chunks at the beginning of the input sequence and individual chunk tails have lesser information due to lack of future context. Moreover, the decoder is autoregressive and errors due to earlier chunks can further affect the predictions downstream. Thus, displaying chunk level outputs as it is can lead to a poor quality translation. One common solution is to use a decoding strategy that selects a partial hypothesis to display instead of the entire output for each input chunk. The subsequent outputs are then conditioned based on the partial hypotheses through forced decoding. By committing to the selected partial hypotheses it can be ensured that the output is stable. There are several decoding strategies that follow different approaches to partial hypothesis selection. We use the local agreement of hypotheses as it balances quality and latency efficiently \cite{liu2020low}.

Local agreement strategy, illustrated with an example in Table~\ref{table:local_agree_example}, is based on the principle that partial hypotheses that remain the same for two consecutive chunk level outputs are more likely to be accurate. A chunk level output is compared to the previous output and the longest common sequence starting from the first token that matches is selected as the partial hypothesis. Naturally, for the first chunk, we do not display any output since there is no previous output to compare to. From the second chunk onwards we display the output based on agreement with the predecessor and use the selected partial hypothesis to condition the future outputs.

\begin{table}[th]
  \caption{Example of chunk-based decoding outputs for the local agreement strategy. The ground-truth transcription for the utterance is “Nature can tell us”}
  \label{table:local_agree_example}
  \small
  \centering
  \adjustbox{max width = \textwidth}{%
\begin{tabular}{@{}llll@{}}
\toprule
\textbf{Input} & \textbf{Displayed} & \textbf{Chunk Output} & \textbf{Agreement} \\ \midrule
(1)            & $\varnothing$      & Nature canned         & $\varnothing$      \\
(1,2)          & $\varnothing$      & Nature can not        & Nature             \\
(1,2,3)        & Nature             & can tell a            & Nature can         \\
(1,2,3,4)      & Nature can         & tell us               & Nature can tell    \\
…              & …                  & …                     & …                  \\ \bottomrule
\end{tabular}}
\end{table}

\subsection{Partial input training}
\label{subsec:partial_input}
Offline ST systems can get efficient at online ST by learning to translate partial input streams. Hence, we create a training corpus with partial inputs using the training data of the offline models. For each training instance, we randomly choose a partial translation ranging from 10\% to 40\% of the total number of tokens starting from the first token. From the audio, we choose an equal proportion of frames (or source text for MT). The ratio is maintained low so that the model learns from partial sequences lacking full input context \cite{liu2020low}. Subsequently, based on the multi-task training scheme described in \cite{niehues2018low}, we generate a dataset with a 1:1 mix of full and partial sequences for all language pairs. The full sequences in the data ensure that the models retain their ability to perform offline ST after the adaptation procedure.

We continue training the offline ST and MT models from the last saved checkpoint using the partial input training corpora. The learning rate is reduced to a quarter of before. The original full sequence mTEDx validation set is used for checkpointing so that the models do not lose their offline ST performance. The final model is the average of the model weights from the last 5 best checkpoints.

\subsection{Latency measurement}
\label{subsec:latency_measurement}
Latency is the measure of time taken for a translation to be generated after the utterance of the speech. Broadly, it is dependent on the engineering design and functioning of the model. We will focus on the latter and examine the latency of a system assuming that the computational resources are equal \cite{liu2020low}.

For an output sequence $w_{1....T}$ the average latency is \begin{equation}
\small
\frac{1}{T} \sum_{t=1}^{T}\left(\text {outputTime}\left(w_{t}\right)-\text { inputTime}\left(w_{t}\right)\right)
\label{eqn:full}
\end{equation}
which can be rewritten as
\begin{equation}
\small
\frac{1}{T} \sum_{t=1}^{T} \text {outputTime}\left(w_{t}\right)-\frac{1}{T} \sum_{t=1}^{T} \text {inputTime}\left(w_{t}\right)
\label{eqn:latency}
\end{equation}
The first term indicates when the translated word $w_{t}$ was generated and the second refers to when the word was uttered (both averaged across all words in the output sequence). Since most ST datasets do not contain a word-level alignment of output translated text to input speech, equation~\ref{eqn:latency} cannot be calculated. Hence, we assume that all output words correspond to the ground truth and drop the second term while calculating latency as it is constant for all systems. While we do not quantify in absolute terms, we can compare the latency of two online ST systems by measuring the difference.

Latency, the first term in equation~\ref{eqn:latency}, can be calculated by measuring the timestamps of output words. In the online decoding framework described previously, input sequences are divided into fixed-size chunks and decoded sequentially. The output text is then displayed at the end of processing each chunk. Therefore, the timestamp for a word can be measured as the chunk index multiplied by the fixed time interval chosen as the chunk size. Finally, it is averaged for all words in the output sequence to give latency in seconds for an online ST system.

\section{Experimental setup}
\label{sec:setup}
\subsection{Data}
\label{subsec:data}
All experiments are performed on the multilingual TEDx corpus for speech recognition and translation \cite{salesky2021multilingual}. Table~\ref{table:mtedx} outlines the corpus statistics of the training data.

\begin{table}[th]
  \caption{Speech transcription and translation data in the mTEDx training set}
  \label{table:mtedx}
  \small
  \centering
  \begin{tabular}{ccccccc}
\hline
\textbf{Source} & \textbf{Transcription}  & \multicolumn{5}{c}{\textbf{Target} (\# utts.)}                                                                           \\
       & (hour, \#utts) & en  & es                       & fr                       & pt                       & it                       \\ \hline
es     & 178, 102k      & 36k & \cellcolor[HTML]{E7E6E6} & 4k                       & 21k                      & 6k                       \\
fr     & 176, 116k      & 30k & 20k                      & \cellcolor[HTML]{E7E6E6} & 13k                      & -                        \\
pt     & 153, 90k       & 31k & -                        & -                        & \cellcolor[HTML]{E7E6E6} & -                        \\
it     & 101, 50k       & -   & -                        & -                        & -                        & \cellcolor[HTML]{E7E6E6} \\ \hline
  \end{tabular}
\end{table}

\subsection{Offline ST models}
\label{subsec:offline_models}
Pre-trained offline ST models used in the experiments are multilingual ST systems from \cite{liu2021maastricht}. All are Transformer \cite{vaswani2017attention} based encoder-decoder models.

We use end-to-end systems jointly trained on ASR and ST data described in Table~\ref{table:mtedx}. We select a model trained on all directions using pseudo-labeled data created by translating ASR transcriptions using an MT system. Another end-to-end model trained only on supervised directions is used for zero-shot experiments. Cascaded systems are built using an end-to-end model as the ASR component followed by an MT model\footnote{Models mentioned have IDs E4, E2, and M1 respectively in \cite{liu2021maastricht}}. The MT model was trained on the data described in Table~\ref{table:mt_train} in both forward and reverse directions.

\begin{table}[th]
  \caption{Text-to-text translation training data from mTEDx corpus across language pairs in number of sentences}
  \label{table:mt_train}
  \small
  \centering
\begin{tabular}{cllccc}
\hline
\multicolumn{1}{l}{} &
  \multicolumn{1}{c}{\textbf{en}} &
  \multicolumn{1}{c}{\textbf{es}} &
  \textbf{fr} &
  \textbf{it} &
  \textbf{pt} \\ \hline
\textbf{en} & \multicolumn{1}{c}{-}    & \multicolumn{1}{c}{36k}  & 30k & 0  & 30k \\
\textbf{es} & \cellcolor[HTML]{E7E6E6} & \multicolumn{1}{c}{-}    & 24k & 6k & 21k \\
\textbf{fr} & \cellcolor[HTML]{E7E6E6} & \cellcolor[HTML]{E7E6E6} & -   & 0  & 13k \\
\textbf{it} &
  \cellcolor[HTML]{E7E6E6} &
  \cellcolor[HTML]{E7E6E6} &
  \multicolumn{1}{l}{\cellcolor[HTML]{E7E6E6}} &
  - &
  0 \\
\textbf{pt} &
  \cellcolor[HTML]{E7E6E6} &
  \cellcolor[HTML]{E7E6E6} &
  \multicolumn{1}{l}{\cellcolor[HTML]{E7E6E6}} &
  \multicolumn{1}{l}{\cellcolor[HTML]{E7E6E6}} &
  - \\ \hline
\end{tabular}
\end{table}

\begin{table*}[ht]
\caption{Translation quality and latency of multilingual cascaded and end-to-end systems averaged for all directions in mTEDx test set}
\label{table:cascade_e2e}
\small
\centering
\adjustbox{max width = \textwidth}{%
\begin{tabular}{@{}lcccccccc@{}}
\toprule
\multicolumn{1}{c}{\multirow{3}{*}{\textbf{Model}}} &
  \multicolumn{4}{c}{\textbf{Translation Quality (BLEU)}} &
  \multicolumn{4}{c}{\textbf{Latency (seconds)}} \\ \cmidrule(l){2-9} 
\multicolumn{1}{c}{} &
  \multicolumn{2}{c}{\textbf{Cascade}} &
  \multicolumn{2}{c}{\textbf{End-to-End}} &
  \multicolumn{2}{c}{\textbf{Cascade}} &
  \multicolumn{2}{c}{\textbf{End-to-End}} \\ \cmidrule(l){2-9} 
\multicolumn{1}{c}{} &
  \textbf{Avg. BLEU} &
  \textbf{$\Delta$ BLEU $\downarrow$} &
  \textbf{Avg. BLEU} &
  \textbf{$\Delta$ BLEU $\downarrow$} &
  \textbf{Avg. Latency} &
  \textbf{$\Delta$ Latency $\uparrow$} &
  \textbf{Avg. Latency} &
  \textbf{$\Delta$ Latency $\uparrow$} \\ \midrule
Offline ST          & 29.70 & 0             & 28.52 & 0             & 9.71 & 0             & 9.73 & 0             \\
Online ST unadapted & 27.83 & 1.87          & 27.28 & 1.24          & 5.65 & 4.06          & 5.64 & 4.09          \\
Online ST adapted   & 28.71 & \textbf{0.99} & 27.95 & \textbf{0.57} & 5.94 & \textbf{3.77} & 5.83 & \textbf{3.90} \\ \bottomrule
\end{tabular}}
\end{table*}

\section{Experiments and results}
\label{sec:experiments}
End-to-end monolingual models can be adapted for online ST using the technique described in Section~\ref{sec:offline_to_online} \cite{liu2020low}. We investigate if it also works for cascaded systems and compare it with end-to-end models. Additionally, we study the impact of the approach across multiple languages. We also analyze the effect of adaptation on zero-shot directions. 

To facilitate online ST, we divide the audio input into 0.5 second chunks before feeding it to the system. BLEU\footnote{sacreBLEU:BLEU+case.mixed+numrefs.1+smooth
.exp+tok.13a+version.1.4.12} scores are used to measure the quality of translation \cite{papineni2002bleu} and latency is quantified as described in Section~\ref{subsec:latency_measurement}.

\subsection{Cascade}
\label{subsec:cascade}
When offline ST systems are used for online ST without adaptation, we see a significant drop in performance in Table~\ref{table:cascade}. Adapting only MT or ASR component leads to no improvement in translation quality as both segments need to learn to handle partial inputs. Adapting both components reduces the relative loss from 1.87 to 0.99 BLEU on online ST. Thus, we are able to reduce the loss in translation quality from an offline to online ST model by 50\% relative. 

\begin{table}[th]
  \caption{Translation quality of multilingual cascaded systems averaged over all directions in the mTEDx test set}
  \label{table:cascade}
  \small
  \centering
  \adjustbox{max width = \textwidth}{%
\begin{tabular}{@{}lcc@{}}
\toprule
\textbf{Cascade Model}    & \textbf{Avg. BLEU} & \textbf{$\Delta$ BLEU $\downarrow$} \\ \midrule
Offline ST        & 29.70         & 0                \\
Online ST unadapted  & 27.83         & 1.87                \\
MT only adapted   & 27.70         & 2.00                \\
ASR only adapted  & 26.47         & 3.23                \\
MT \& ASR adapted & 28.71         & \textbf{0.99}                \\ \bottomrule
\end{tabular}}
\end{table}

\subsection{Cascade vs. end-to-end}
\label{subsec:cascade_e2e}
We compare the performance of ASR and MT adapted cascaded system with an end-to-end model adapted with partial ASR and ST data in all directions in Table~\ref{table:cascade_e2e}. We see that the adaptation in both architectures improves the translation quality and reduces the loss due to online translation by 50\% relative. However, the loss in translation quality relative to offline ST model of 0.99 BLEU for the cascaded system is greater compared to 0.57 BLEU for the end-to-end model. One reason could be that there is error propagation between ASR and MT components when learning to translate partial inputs whereas the end-to-end model is able to learn better language representations. 

The gains in latency though are very similar, 3.77 seconds and 3.90 seconds for cascaded and end-to-end models respectively. It is likely that improvement in latency is independent of loss or gain in translation quality and choice of architecture.  

\subsection{Multilingual translation}
\label{subsec:multilingual}
We analyze the effect of adaptation in cascaded and end-to-end models for all languages in Table~\ref{table:multilingual}. Relative gains in latency for the adapted models are not only similar for both architectures but are also consistent across languages. Most translation directions gain around 40\% in latency compared to offline systems. 

On the other hand, translation quality is different for various languages as observed by the inconsistency of relative loss in BLEU. End-to-end models have similar or lower loss than cascaded systems for all language pairs except es-fr. In Table~\ref{table:mtedx} we see that es-fr has the least ST training data which hampers the translation for the end-to-end system. However, the cascaded system has the least relative loss for es-fr among all directions due to a large number of es-es ASR transcriptions and es-fr MT translations available as seen in Tables~\ref{table:mtedx} and~\ref{table:mt_train} respectively. 

Similar trends can be observed for it-en and it-es. Both directions have a high relative loss in BLEU for the cascaded system as it-it has the least amount of ASR transcriptions in Table~\ref{table:mtedx}, it-en MT translations are not available and it-es has the least training data in Table~\ref{table:mt_train}. On the contrary, end-to-end systems have the least relative loss in BLEU for it-en and it-es. They greatly benefit from training with pseudo labeled data created by translating it-it ASR transcriptions using the MT system specifically since no ST data was available for Italian audio. Therefore, it is evident that the quality of translation is resource-dependent while latency is not. 

\begin{table*}[ht]
\caption{Translation quality and latency for multilingual cascaded and end-to-end systems across all languages in the mTEDx test set}
\label{table:multilingual}
\small
\centering
\adjustbox{max width = \textwidth}{%
\begin{tabular}{@{}cclccccccccccc@{}}
\toprule
\multicolumn{3}{c}{\textbf{Description}} &
  \textbf{es-en} &
  \textbf{es-fr} &
  \textbf{es-pt} &
  \textbf{fr-en} &
  \textbf{fr-es} &
  \textbf{fr-pt} &
  \textbf{pt-en} &
  \textbf{pt-es} &
  \textbf{it-en} &
  \textbf{it-es} &
  \textbf{Avg.} \\ \midrule
  \multirow{6}{*}{\rotatebox[origin=c]{90}{Cascade}} &
  \multirow{3}{*}{\begin{tabular}[c]{@{}c@{}}Quality\\ (BLEU)\end{tabular}} &
  Offline &
  26.6 &
  26.9 &
  36.2 &
  30.3 &
  33.7 &
  33.5 &
  30.3 &
  34.2 &
  18.5 &
  26.8 &
  29.70 \\
 &
   &
  Adapted &
  25.6 &
  26.5 &
  35.0 &
  29.3 &
  32.9 &
  32.5 &
  28.8 &
  33.3 &
  17.6 &
  25.6 &
  28.71 \\
 &
   &
  \% Loss $\downarrow$ &
  \textbf{3.8\%} &
  \textbf{1.5\%} &
  \textbf{3.3\%} &
  \textbf{3.3\%} &
  \textbf{2.4\%} &
  \textbf{3.0\%} &
  \textbf{5.0\%} &
  \textbf{2.6\%} &
  \textbf{4.9\%} &
  \textbf{4.5\%} &
  \textbf{3.3\%} \\ \cmidrule(l){2-14} 
 &
  \multirow{3}{*}{\begin{tabular}[c]{@{}c@{}}Latency\\ (seconds)\end{tabular}} &
  Offline &
  10.24 &
  10.25 &
  10.39 &
  8.39 &
  8.47 &
  8.42 &
  10.28 &
  10.30 &
  10.15 &
  10.21 &
  9.71 \\
 &
   &
  Adapted &
  6.18 &
  6.10 &
  6.22 &
  5.25 &
  5.37 &
  5.39 &
  6.19 &
  6.24 &
  6.22 &
  6.22 &
  5.94 \\
 &
   &
  \% Gain $\uparrow$ &
  \textbf{39.6\%} &
  \textbf{40.5\%} &
  \textbf{40.1\%} &
  \textbf{37.4\%} &
  \textbf{36.6\%} &
  \textbf{36.0\%} &
  \textbf{39.8\%} &
  \textbf{39.4\%} &
  \textbf{38.7\%} &
  \textbf{39.1\%} &
  \textbf{38.8\%} \\ \midrule
  \multirow{6}{*}{\rotatebox[origin=c]{90}{End-to-end}} &
  \multirow{3}{*}{\begin{tabular}[c]{@{}c@{}}Quality\\ (BLEU)\end{tabular}} &
  Offline &
  25.0 &
  20.8 &
  33.8 &
  30.0 &
  33.3 &
  30.5 &
  28.4 &
  34.2 &
  20.4 &
  28.8 &
  28.52 \\
 &
   &
  Adapted &
  24.0 &
  19.5 &
  33.2 &
  29.2 &
  32.8 &
  30.4 &
  27.6 &
  33.9 &
  20.1 &
  28.8 &
  27.95 \\
 &
   &
  \% Loss $\downarrow$ &
  \textbf{4.0\%} &
  \textbf{6.3\%} &
  \textbf{1.8\%} &
  \textbf{2.7\%} &
  \textbf{1.5\%} &
  \textbf{0.3\%} &
  \textbf{2.8\%} &
  \textbf{0.9\%} &
  \textbf{1.5\%} &
  \textbf{0.0\%} &
  \textbf{2.0\%} \\ \cmidrule(l){2-14} 
 &
  \multirow{3}{*}{\begin{tabular}[c]{@{}c@{}}Latency\\ (seconds)\end{tabular}} &
  Offline &
  10.26 &
  10.01 &
  10.37 &
  8.49 &
  8.56 &
  8.51 &
  10.56 &
  10.36 &
  10.05 &
  10.14 &
  9.73 \\
 &
   &
  Adapted &
  6.00 &
  6.19 &
  6.09 &
  5.16 &
  5.25 &
  5.33 &
  6.04 &
  6.17 &
  6.01 &
  6.09 &
  5.83 \\
 &
   &
  \% Gain $\uparrow$ &
  \textbf{41.5\%} &
  \textbf{38.1\%} &
  \textbf{41.3\%} &
  \textbf{39.2\%} &
  \textbf{38.6\%} &
  \textbf{37.4\%} &
  \textbf{42.8\%} &
  \textbf{40.4\%} &
  \textbf{40.2\%} &
  \textbf{39.9\%} &
  \textbf{40.1\%} \\ \bottomrule
\end{tabular}}
\end{table*}

\subsection{Zero-shot translation}
\label{subsec:zero-shot}
Multilingual models can translate in language directions that do not exist in the training data. So we investigate the impact of the adaptation procedure on online zero-shot ST in Table~\ref{table:zero-shot}. We compare the performance of the model trained on all directions to the model trained only on supervised directions. First, we observe that the adapted models have similar relative gains in average BLEU compared to the unadapted models in the supervised direction for both architectures. This reconfirms the efficacy of the technique. Interestingly, gains in average BLEU in zero-shot directions for the model trained only on supervised directions is comparable to the model trained on all directions for cascade and end-to-end systems. The knowledge gained in translating partial sentences is transferred to zero-shot directions while performing online ST. 

\begin{table*}[ht]
\caption{Translation quality of multilingual models trained on all language directions vs. trained only on supervised directions in BLEU averaged over supervised and zero-shot directions in the mTEDx test set}
\label{table:zero-shot}
\small
\centering
\adjustbox{max width = \textwidth}{%
\begin{tabular}{@{}clcccc@{}}
\toprule
\multicolumn{2}{c}{\multirow{2}{*}{\textbf{Description}}} &
  \multicolumn{2}{c}{\textbf{Supervised Directions (Avg. BLEU)}} &
  \multicolumn{2}{c}{\textbf{Zero-shot Directions (Avg. BLEU)}} \\ \cmidrule(l){3-6} 
\multicolumn{2}{c}{} &
  \multicolumn{1}{l}{\textbf{Trained on All Directions}} &
  \multicolumn{1}{l}{\textbf{Trained on Supervised Only}} &
  \multicolumn{1}{l}{\textbf{Trained on All Directions}} &
  \multicolumn{1}{l}{\textbf{Trained on Supervised Only}} \\ \midrule
\multirow{3}{*}{\rotatebox[origin=c]{90}{Cascade}}    & Online ST unadapted & 29.07 & 28.43 & 24.93         & 23.87         \\
                            & Online ST adapted   & 30.09 & 29.39 & 25.50         & 24.33         \\
                            & $\Delta$ BLEU $\uparrow$          & 1.01  & 0.96  & \textbf{0.57} & \textbf{0.47} \\ \midrule
\multirow{3}{*}{\rotatebox[origin=c]{90}{E2E}} & Online ST unadapted & 27.69 & 23.10 & 26.33         & 13.83         \\
                            & Online ST adapted   & 28.10 & 23.57 & 27.60         & 14.83         \\
                            & $\Delta$ BLEU $\uparrow$          & 0.41  & 0.47  & \textbf{1.27} & \textbf{1.00} \\ \bottomrule
\end{tabular}}
\end{table*}

\section{Conclusion}
\label{sec:conclusion}
Multilingual simultaneous speech translation applications have the potential to reduce communication barriers by enabling a diverse population to follow live discussions and programs in real-time. Training an offline ST system with partial training data followed by implementing a local agreement decoding strategy is one approach to building such a system. It is capable of handling quality latency trade-off for a monolingual end-to-end system and we demonstrated the effectiveness of the method for online ST in cascaded systems. Additionally, we showed that cascaded systems had a larger relative loss in translation quality upon adaptation than end-to-end systems while gains in latency were similar. We analyzed the effect of the approach on multilingual data and highlighted the impact of training resources on translation quality in cascaded and end-to-end systems. Furthermore, we demonstrated that the adaptation technique leads to gains in translation quality in zero-shot directions not seen in training. 

Our findings suggest that an effective multilingual online ST system can be built using the adaptation procedure. Importantly, the technique is simple to implement. It is model agnostic, does not require external data curated for online ST, and by leveraging a pre-trained model it is time and resource efficient. Moreover, we can ensure that the final adapted model retains its offline ST performance. We hope that our research will contribute to the development of hybrid multilingual offline and low-latency online ST models for practical applications.

\section{Acknowledgements}
We would like to thank Danni Liu for her vital support and timely inputs throughout the research.

\bibliographystyle{IEEEtran}
\bibliography{msst}

\begin{thebibliography}{10}
\providecommand{\url}[1]{#1}
\csname url@samestyle\endcsname
\providecommand{\newblock}{\relax}
\providecommand{\bibinfo}[2]{#2}
\providecommand{\BIBentrySTDinterwordspacing}{\spaceskip=0pt\relax}
\providecommand{\BIBentryALTinterwordstretchfactor}{4}
\providecommand{\BIBentryALTinterwordspacing}{\spaceskip=\fontdimen2\font plus
\BIBentryALTinterwordstretchfactor\fontdimen3\font minus
  \fontdimen4\font\relax}
\providecommand{\BIBforeignlanguage}[2]{{%
\expandafter\ifx\csname l@#1\endcsname\relax
\typeout{** WARNING: IEEEtran.bst: No hyphenation pattern has been}%
\typeout{** loaded for the language `#1'. Using the pattern for}%
\typeout{** the default language instead.}%
\else
\language=\csname l@#1\endcsname
\fi
#2}}
\providecommand{\BIBdecl}{\relax}
\BIBdecl

\bibitem{pham2019very}
N.-Q. Pham, T.-S. Nguyen, J.~Niehues, M.~M{\"u}ller, S.~St{\"u}ker, and
  A.~Waibel, ``Very deep self-attention networks for end-to-end speech
  recognition,'' \emph{arXiv preprint arXiv:1904.13377}, 2019.

\bibitem{ansari2020findings}
E.~Ansari, A.~Axelrod, N.~Bach, O.~Bojar, R.~Cattoni, F.~Dalvi, N.~Durrani,
  M.~Federico, C.~Federmann, J.~Gu \emph{et~al.}, ``Findings of the iwslt 2020
  evaluation campaign,'' in \emph{Proceedings of the 17th International
  Conference on Spoken Language Translation}, 2020, pp. 1--34.

\bibitem{zhao2021volctrans}
C.~Zhao, Z.~Liu, J.~Tong, T.~Wang, M.~Wang, R.~Ye, Q.~Dong, J.~Cao, and L.~Li,
  ``The volctrans neural speech translation system for iwslt 2021,''
  \emph{arXiv preprint arXiv:2105.07319}, 2021.

\bibitem{li2020multilingual}
X.~Li, C.~Wang, Y.~Tang, C.~Tran, Y.~Tang, J.~Pino, A.~Baevski, A.~Conneau, and
  M.~Auli, ``Multilingual speech translation with efficient finetuning of
  pretrained models,'' \emph{arXiv preprint arXiv:2010.12829}, 2020.

\bibitem{niehues2018low}
J.~Niehues, N.-Q. Pham, T.-L. Ha, M.~Sperber, and A.~Waibel, ``Low-latency
  neural speech translation,'' \emph{arXiv preprint arXiv:1808.00491}, 2018.

\bibitem{liu2020low}
D.~Liu, G.~Spanakis, and J.~Niehues, ``Low-latency sequence-to-sequence speech
  recognition and translation by partial hypothesis selection,'' \emph{arXiv
  preprint arXiv:2005.11185}, 2020.

\bibitem{dessloch2018kit}
F.~Dessloch, T.-L. Ha, M.~M{\"u}ller, J.~Niehues, T.-S. Nguyen, N.-Q. Pham,
  E.~Salesky, M.~Sperber, S.~St{\"u}ker, T.~Zenkel \emph{et~al.}, ``Kit lecture
  translator: Multilingual speech translation with one-shot learning,'' in
  \emph{Proceedings of the 27th International Conference on Computational
  Linguistics: System Demonstrations}, 2018, pp. 89--93.

\bibitem{inaguma2019multilingual}
H.~Inaguma, K.~Duh, T.~Kawahara, and S.~Watanabe, ``Multilingual end-to-end
  speech translation,'' in \emph{2019 IEEE Automatic Speech Recognition and
  Understanding Workshop (ASRU)}.\hskip 1em plus 0.5em minus 0.4em\relax IEEE,
  2019, pp. 570--577.

\bibitem{zeng2021multilingual}
X.~Zeng, L.~Li, and Q.~Liu, ``Multilingual speech translation with unified
  transformer: Huawei noah's ark lab at iwslt 2021,'' \emph{arXiv preprint
  arXiv:2106.00197}, 2021.

\bibitem{fujita2013simple}
T.~Fujita, G.~Neubig, S.~Sakti, T.~Toda, and S.~Nakamura, ``Simple, lexicalized
  choice of translation timing for simultaneous speech translation.'' in
  \emph{INTERSPEECH}, 2013, pp. 3487--3491.

\bibitem{sridhar2013segmentation}
V.~K.~R. Sridhar, J.~Chen, S.~Bangalore, A.~Ljolje, and R.~Chengalvarayan,
  ``Segmentation strategies for streaming speech translation,'' in
  \emph{Proceedings of the 2013 Conference of the North American Chapter of the
  Association for Computational Linguistics: Human Language Technologies},
  2013, pp. 230--238.

\bibitem{gu2016learning}
J.~Gu, G.~Neubig, K.~Cho, and V.~O. Li, ``Learning to translate in real-time
  with neural machine translation,'' \emph{arXiv preprint arXiv:1610.00388},
  2016.

\bibitem{arivazhagan2019monotonic}
N.~Arivazhagan, C.~Cherry, W.~Macherey, C.-C. Chiu, S.~Yavuz, R.~Pang, W.~Li,
  and C.~Raffel, ``Monotonic infinite lookback attention for simultaneous
  machine translation,'' \emph{arXiv preprint arXiv:1906.05218}, 2019.

\bibitem{liu2021ustc}
D.~Liu, M.~Du, X.~Li, Y.~Hu, and L.~Dai, ``The ustc-nelslip systems for
  simultaneous speech translation task at iwslt 2021,'' \emph{arXiv preprint
  arXiv:2107.00279}, 2021.

\bibitem{cho2016can}
K.~Cho and M.~Esipova, ``Can neural machine translation do simultaneous
  translation?'' \emph{arXiv preprint arXiv:1606.02012}, 2016.

\bibitem{ma2018stacl}
M.~Ma, L.~Huang, H.~Xiong, R.~Zheng, K.~Liu, B.~Zheng, C.~Zhang, Z.~He, H.~Liu,
  X.~Li \emph{et~al.}, ``Stacl: Simultaneous translation with implicit
  anticipation and controllable latency using prefix-to-prefix framework,''
  \emph{arXiv preprint arXiv:1810.08398}, 2018.

\bibitem{salesky2021multilingual}
E.~Salesky, M.~Wiesner, J.~Bremerman, R.~Cattoni, M.~Negri, M.~Turchi, D.~W.
  Oard, and M.~Post, ``The multilingual tedx corpus for speech recognition and
  translation,'' \emph{arXiv preprint arXiv:2102.01757}, 2021.

\bibitem{liu2021maastricht}
D.~Liu and J.~Niehues, ``Maastricht university’s multilingual speech
  translation system for iwslt 2021,'' in \emph{Proceedings of the 18th
  International Conference on Spoken Language Translation (IWSLT 2021)}, 2021,
  pp. 138--143.

\bibitem{vaswani2017attention}
A.~Vaswani, N.~Shazeer, N.~Parmar, J.~Uszkoreit, L.~Jones, A.~N. Gomez,
  {\L}.~Kaiser, and I.~Polosukhin, ``Attention is all you need,''
  \emph{Advances in neural information processing systems}, vol.~30, 2017.

\bibitem{papineni2002bleu}
K.~Papineni, S.~Roukos, T.~Ward, and W.-J. Zhu, ``Bleu: a method for automatic
  evaluation of machine translation,'' in \emph{Proceedings of the 40th annual
  meeting of the Association for Computational Linguistics}, 2002, pp.
  311--318.

\end{thebibliography}

\end{document}